\documentclass{article}

\usepackage{nips_syn4ML}

\usepackage[utf8]{inputenc} 
\usepackage[T1]{fontenc}    
\usepackage{hyperref}       
\usepackage{url}            
\usepackage{booktabs}       
\usepackage{amsfonts}       
\usepackage{nicefrac}       
\usepackage{microtype}      
\usepackage{xcolor}         
\usepackage{graphicx}
\usepackage{amsmath}
\usepackage{amssymb}
\usepackage{mathtools}
\usepackage{amsthm}
\usepackage{graphics, setspace, calc}
\usepackage{abstract}
\usepackage{enumitem}
\usepackage{paralist}
\usepackage{bbm}
\usepackage{color}
\usepackage{caption}
\usepackage{algpseudocode}
\usepackage{lipsum} 
\usepackage[makeroom]{cancel}
\usepackage{natbib}
\setcitestyle{numbers}
\setcitestyle{square}
\usepackage{wrapfig}
\usepackage[math]{alp}
\usepackage{algorithm}

\title{EDGE++: Improved Training and Sampling of EDGE}

\begin{document}

\maketitle
\def\thefootnote{*}\footnotetext{These authors contributed equally to this work.}\def\thefootnote{\arabic{footnote}}

\begin{abstract}
Recently developed deep neural models like NetGAN, CELL, and Variational Graph Autoencoders have made progress but face limitations in replicating key graph statistics on generating large graphs. Diffusion-based methods have emerged as promising alternatives, however, most of them present challenges in computational efficiency and generative performance. EDGE is effective at modeling large networks, but its current denoising approach can be inefficient, often leading to wasted computational resources and potential mismatches in its generation process. In this paper, we propose enhancements to the EDGE model to address these issues. Specifically, we introduce a degree-specific noise schedule that optimizes the number of active nodes at each timestep, significantly reducing memory consumption. Additionally, we present an improved sampling scheme that fine-tunes the generative process, allowing for better control over the similarity between the synthesized and the true network. Our experimental results demonstrate that the proposed modifications not only improve the efficiency but also enhance the accuracy of the generated graphs, offering a robust and scalable solution for graph generation tasks.
\end{abstract}

\section{Introduction}

The generation of large graphs has been accomplished using random graph models~ \citep{newman2002random}, such as the Stochastic-Block Model~(SBM)~\citep{holland1983stochastic}. Despite their use, these models fall short in capturing complex structures, paving the way to the development of deep neural models. Recently, several neural methods, including NetGAN~\cite{bojchevski2018netgan}, CELL~\citep{rendsburg2020netgan}, and Variational Graph Autoencoders~\citep{kipf2016variational}, have been proposed to model large graphs. However, \citet{chanpuriya2021power} points out that they are edge-independent model and are still incapable of reproducing key statistics unless they memorize the training graph, i.e., high edge overlap between the generated graphs and the original one. An edge-independent model generates all edges independently at once. In constrast, edge-dependent models like diffusion-based graph models~\citep{sohl2015deep,ho2020denoising} and autoregressive graph models~\citep{you2018graphrnn, liao2019efficient}, have shown promise with small graphs~\citep{jo2022score,vignac2022digress}. In particular, diffusion-based graph models demonstrate better modeling capbility as they avoid the long-term memory issues. 

While demonstrating significant success, the majority of existing diffusion models fail to generate large networks with thousands of nodes. Recently, \citet{chen2023efficient} advocates to diffuse a graph into an empty graph and leverages a neural network to reverse the edge-removal process (see Fig.~\ref{fig:edge-overview}(a)). In the edge-removal process, it identifies that not all nodes participate in the edge-formation process in every timestep, and proposes to first select the active nodes then predict edge formation only among them. Furthermore, it shows that one can use a prescribed degree sequence to guide the graph generation. This denoising scheme notably reduces both the computational cost and task complexity.

While EDGE has demonstrated powerful capability in modeling large networks, the computational power of the denoising network is not fully utilized. As demonstrated in Fig.~\ref{fig:edge-overview}(b), by using default noise schedule in vanilla diffusion models, the computation is highly concentrated in the second half of the denoising process, wasting the network capability in the first half. Such an uneven distribution of the number of active nodes also leads to a higher memory consumption, as the space complexity is upperbounded by the 
largest number. Moreover, while using degree-guided generation, we observe that there is a mismatch between the diffusion and denoising processes (see Fig.~\ref{fig:edge-overview}(b,c)). Such inconsistency may potentially deteriorate the generative performance.

In this work, we propose two components to address the aforementioned issues in EDGE. First, we propose a degree-specific noise schedule to control the number of active nodes at each timestep. The proposed noise schedule significantly reduces the largest number of active nodes during denoising, saving much more computation in terms of memory. Second, we propose an improved sampling scheme, which fixes the generation error potentially made in each denoising step. Experiment results show that by adopting the proposed techniques, we obtain significant memory savings in training EDGE, and achieve better generative performance in Polblogs and PPI datasets. Moreover, we showcase how to perform graph generation with EO control by leveraging the proposed techniques.

\begin{figure}[t]
    \centering
    \begin{tabular}{cc}
\multicolumn{2}{c}{\includegraphics[width=0.8\textwidth]{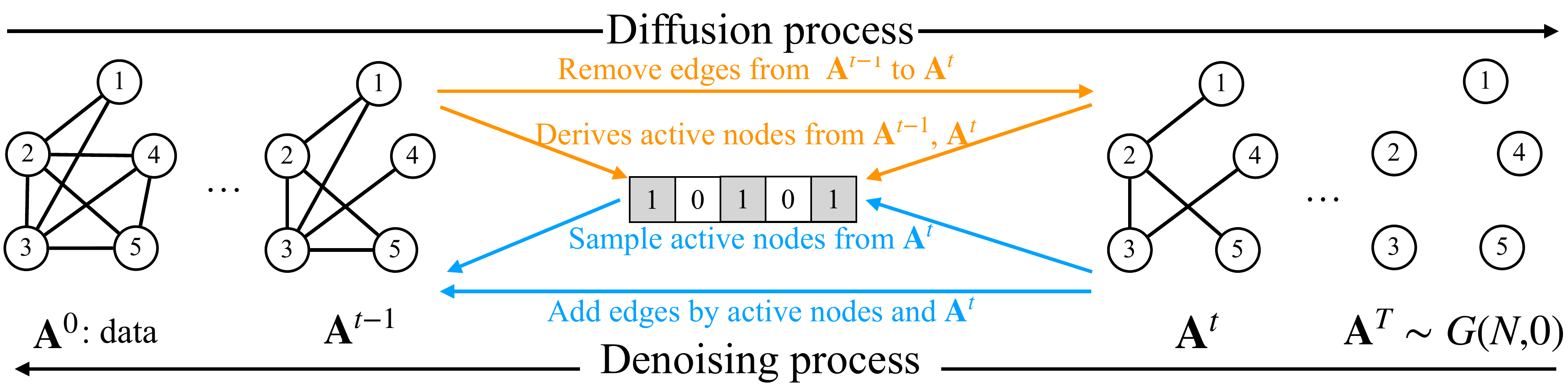}}\\
\multicolumn{2}{c}{(a) Overview of the EDGE framework}\\
\includegraphics[width=0.42\textwidth]{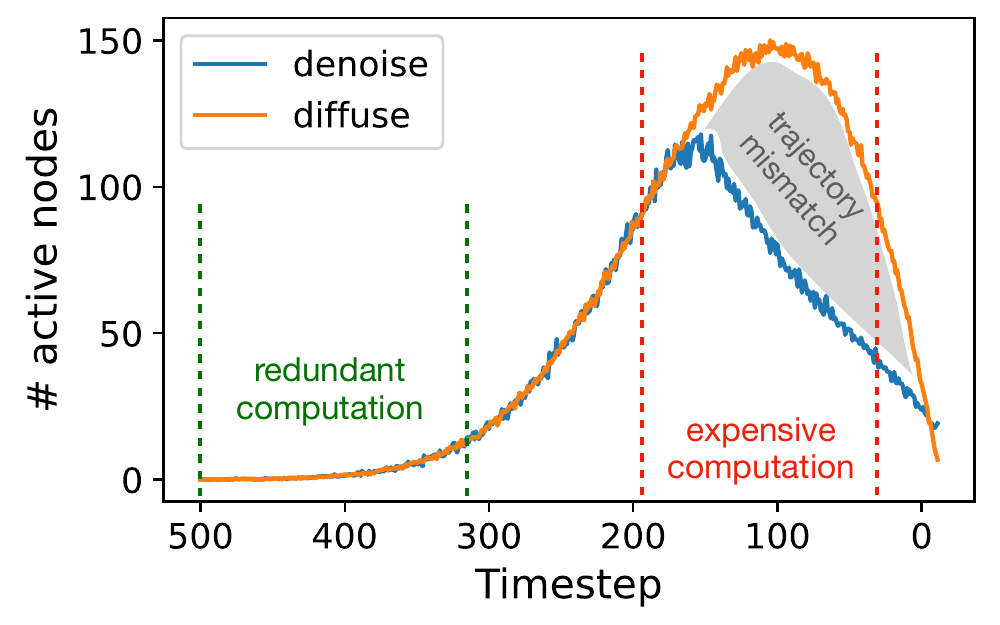}&\includegraphics[width=0.42\textwidth]{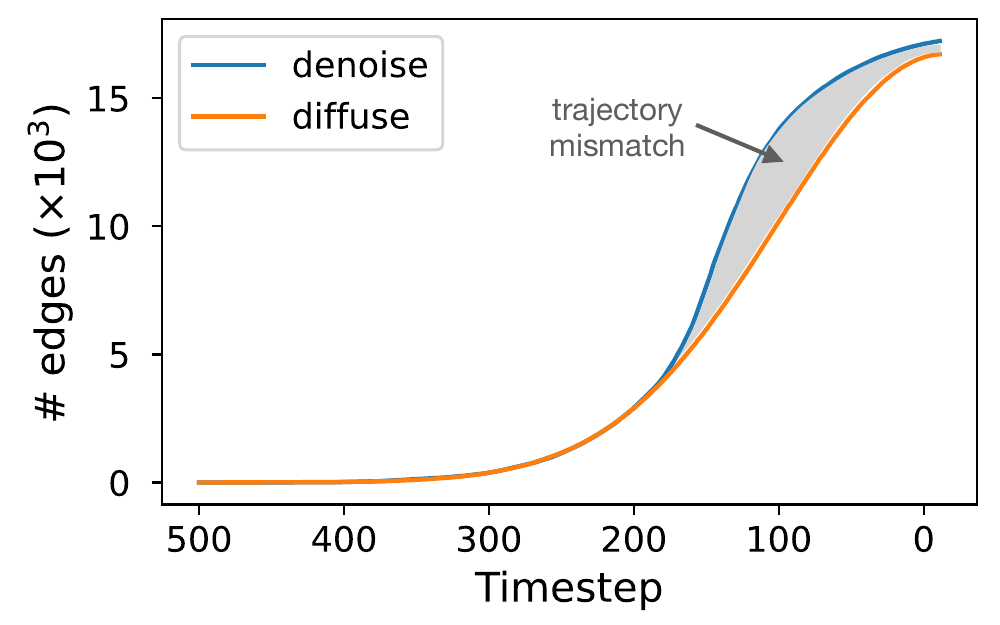}\\
(b) Behavior of active nodes in EDGE & (c) Behavior of generated edges in EDGE
\end{tabular}
\vspace{-0.2em}
\caption{Figure 1. (a) EDGE defines an edge-removal process and learns a model to reverse it. In the denoising process, it first identifies nodes that may have edges added (active node), then predicts edges between them. (b, c) Denoising and diffusion processes of active nodes and edges on Polblogs. The denoising process in EDGE is inconsistent with the reverse of the diffusion process. This mismatch issue appears in the active nodes prediction and consequently leads to unsatisfied generative performance.}
\label{fig:edge-overview}
\vspace{-1.2em}
\end{figure}
\vspace{-1em}
\section{Background}
\vspace{-0.8em}
We are interested in a graph generative model that considers the following diffusion process over variables $\bA^{1:T}$:
\vspace{-0.6em}
\begin{align}
    q(\bA^{1:T}|\bA^0) = \prod_{t=1}^Tq(\bA^t|\bA^{t-1}),~\text{where}~q(\bA^t|\bA^{t-1}) &= \prod_{i,j: i<j} \calB(\bA^t_{i,j}; (1-\beta_t) \bA^{t-1}_{i,j}+\beta_tp).\nonumber
\end{align}
\vspace{-1.1em}

Here $\bA^0$ is the data adjacency matrix, and $\calB(x; \mu)$ is the Bernoulli distribution over variable $x$ with probability $\mu$. And $\beta_t$ is the noise schedule parameter at timestep $t$. Following the convention, we define $\alpha_t\!=\!1\!-\!\beta_t$ and $\bar{\alpha}_t\!=\!\prod_{\tau=1}^t\alpha_\tau$. Ideally, $\beta_{1:T}$ are tuned such that the marginal $q(\bA^T|\bA^0)$ converges to some prior distribution $p(\bA^T)$. Specifically, $p(\bA^T)$ is an \erdosrenyi graph model $G(N, p)$, with $N$ being the graph size and $p$ being the probability an edge may occur between any two nodes. EDGE~\citep{chen2023efficient} set $p=0$ and simplify the diffusion process into an edge-removal process (Fig.~\ref{fig:edge-overview}(a)). Under this setting, it proposes two techniques to reduce the computation cost and learning complexity.

\vspace{-0.8em}
\paragraph{Introducing the active node variables.} Due to the sparsity property of a graph, EDGE identifies that only a portion of the nodes may have their edges removed at every timestep, and derives the following equivalent diffusion process:
\vspace{-0.6em}
\begin{align}
q(\bA^{1:T}|\bA^0) = q(\bA^{1:T}, \bs^{1:T}|\bA^0) = \prod_{t=1}^{T}q(\bA^t|\bA^{t-1})q(\bs^t|\bA^{t-1}, \bA^{t}).\nonumber
\end{align}
This is because $\bs^t$ is deterministic given $\bA^{t-1}$ and $\bA^{t}$. In the denoising process, for each timestep, one can first decide which nodes will be active given $\bA^t$, then only perform edge prediction among the active nodes. The denoising process can be formulated as 
\begin{align}
p_\theta(\bA^{0:T},\bs^{1:T})=p(\bA^T)\prod_{t=1}^{T} p_\theta(\bA^{t-1}|\bA^t,\bs^t)p_\theta(\bs^t|\bA^t),\nonumber
\end{align}
then the learning objective is to maximize the variational lower bound $\calL(\bA^0;\theta)$ of $\log{p_\theta(\bA^0)}$:
\vspace{-0.5em}
\begin{align}
\!\!\mathbb{E}_q\Bigg[\!\!\log{\frac{p(\bA^T)}
{q(\bA^T|\bA^0)}}\!+\!\log{p_\theta(\bA^0|\bA^1\!,\bs^1)}\!+\!\sum_{t=2}^T\log{\frac{p_\theta(\bA^{t-1}|\bA^t,\bs^t)}{q(\bA^{t-1}|\bA^t,\bs^t,\bA^0)}}\!+\!\sum_{t=1}^T\log{\frac{p_\theta(\bs^t|\bA^t)}{q(\bs^t|\bA^t,\bA^0)}}\Bigg].\nonumber
\end{align}
\paragraph{Degree-guided graph generation.} EDGE further shows that if the initial degree $\bd^0$ of the generated graph is given, one doesn't need to learn the active node predictor $p_\theta(\bs^t|\bA^t)$. Specifically, it shows that given the initial degree $\bd^0$ and the current degree $\bd^t$, the active node posterior is
\begin{align}
     &q(\bs^t|\bd^t, \bd^0)=\prod^N_{i=1}q(\bs^t_i|\bd^t_i, \bd^0_i),~\text{where}~q(\bs^t_i|\bd^t_i, \bd^0_i)=\calB\bigg(\bs^t_i;1-\Big(1-\frac{\beta_{t}\bar{\alpha}_{t-1}}{1-\bar{\alpha}_{t}}\Big)^{\bd^0_i\!-\!\bd^t_i}\bigg). \label{eq:active-post}
\end{align}
To sample a graph, one needs to sample a degree sequence $\bd^0$ first, then replace the parameterized active node distribution $p_\theta(\bs^t|\bA^t)$ with $q(\bs^t|\bd^t, \bd^0)$. The degree-guided generation can significantly reduce the model learning complexity, greatly improving the generation accuracy.

Since EDGE defines latent variables on two levels of granularities -- nodes and edges, the behavior of the active nodes should also be taken into consideration. Fig.~\ref{fig:edge-overview}(b) visualizes the node behavior when defining linear noise schedule on edges, the number of active nodes varies unevenly over timesteps, making the modeling of some timesteps wasteful and the sampling computation redundant. Moreover, we observe that there is a mismatch issue during sampling, making the generated graphs having a higher volume than the ground-truth graph (Fig~\ref{fig:edge-overview}(c)). Next, we present two techniques to address those limitations without modifying the framework of EDGE.


\vspace{-0.5em}\section{Methodologies}\vspace{-0.5em}

In this section, we first elaborate on how to improve the noise schedule based on active node control, and then we present volume-preserved sampling, which alleviates the aforementioned mismatching issue. 

\vspace{-0.5em}
\subsection{Improved noise schedule}\vspace{-0.5em}
\label{sec:3.1}
While EDGE uses existing noise schedule schemes, which are defined on edges, we argue that a better schedule principle should focus on the nodes. Denote $\gamma_{1:T}$ to be an active node schedule, where $\gamma_t>0$ for all $t$ is an unnormalized portion of the total number of nodes. The reason it is unnormalized is that the actual number of active nodes is data-specific. We introduce a free parameter $K$ such that $g_{\gamma_t}(K):=KN\gamma_t$ is the actual number of active nodes, we defer the discussion of the use of $K$ later in this section.

Given $\gamma_{1:T}$, the goal is to find the corresponding edge noise schedule $\alpha_{1:T}$. In the following, we will first draw the connection between the edge noise schedule and the expected number of active nodes for each timestep, then we show how to obtain the parameters $\alpha_{1:T}$ that satisfy the given active node schedule.

\vspace{-0.5em}
\paragraph{Connecting edge noise schedule to active node control.} Since the active node schedule is data-specific, let $\bd^0\in\mathbb{N}^N$ be the degree sequence of a graph, given $\alpha_{1:T}$, the expected number of active nodes $h_{\bd^0}(\alpha_{1:t},t)\in\mathbb{R}$ at timestep $t$ is
\vspace{-0.8em}
\begin{align}
     h_{\bd^0}(\alpha_{1:t},t)=
     \begin{cases}
        \displaystyle\sum_{i=1}^{N} (1-\alpha_1^{\bd_i^0}),~&\text{if}~t=1\\
        \displaystyle\sum_{i=1}^{N} \sum_{\bd_i^{t-1}=1}^{\bd_i^0}(1-\alpha_t^{\bd_i^{t-1}}) \ 
        \mathrm{Bin}(k=\bd_i^{t-1}, n=\bd_i^0, p=\bar{\alpha}_{t-1}),&\text{otherwise}\label{eq:expect-num-active}
     \end{cases}.
\end{align}
Here $\mathrm{Bin}(k,n,p)$ is a binomial distribution parameterized by number of trails $n$ and probability $p$. Intuitively, the expected number of active nodes is computed as $\sum_{i=1}^N q(\bs^t_i=1|\bd_i^0)$, where $ q(\bs^t_i|\bd_i^0)$ is the distribution of active node $i$ at timestep $t$. When $t>1$, $q(\bs^t_i|\bd_i^0)$ is computed by the marginalization
\vspace{-1.em}
\begin{align}
    q(\bs^t_i|\bd_i^0) = \sum_{\bd_i^{t-1}=1}^{\bd_i^0}q(\bs^t_i|\bd_i^{t-1})q(\bd_i^{t-1}|\bd_i^0),
\end{align}
where both $q(\bs^t_i|\bd_i^{t-1})$ and $q(\bd_i^{t-1}|\bd_i^0)$ can be expressed analytically~\citep{chen2023efficient}. We further provide a detailed derivation in the App.~\ref{app:derivation-connect}.
\vspace{-0.8em}
\paragraph{Finding the corresponding edge noise schedule $\alpha_{1:T}^*$.} With the shown relation between $\alpha_{1:T}$ and active node behavior, we now can get $\alpha_{1:T}^*$ for any active node schedule. Specifically, we can obtain $\alpha_{1:T}^*$ by solving
\vspace{-0.5em}
\begin{align}
    \alpha_{1:T}^* =\argmin_{\alpha_{1:T}} \underbrace{\sum_{t=1}^T 
    \Big(h_{\bd^0}(\alpha_{1:t},t) -g_{\gamma_t}(K)\Big)^2}_{\calL(\alpha_{1:T};K,\gamma_{1:T},\bd^0)}, ~\text{s.t.}~\prod_{t=1}^T\alpha_t\approx0.\label{eq:alpha-obj}
\end{align}\vspace{-0.2em}
Here we are actually optimizing $\alpha_{1:T}$ such that the expected number of active nodes matches the desired one (i.e., $g_{\gamma_t}(K)$) at each timestep. Recall that we need the marginal distribution $q(\bA^T|\bA^0)$ converge to $G(N,0)$, so $\prod\alpha_t$ should converge to 0. This constraint is imposed when solving the objective.

To solve for $\alpha_{1:T}^*$ that satisfies the constraint, we introduce the parameter $K$, which is solved along with $\alpha_{1:T}$. The parameter $K$ is tuned such that (1) the objective loss is sufficiently low (when $K$ is small enough); (2) and the constraint $\prod\alpha_t\approx 0$ is satisfied (when $K$ is large enough). In practice, we solve $K$ and $\alpha_{1:T}^*$ alternatively via binary search (See Alg.~\setcounter{figure}{0}\ref{alg-obj}). Given specific $K$, we solve the objective using a numerical solver.


\subsection{Volume-preserved Sampling}

\begin{wrapfigure}[16]{r}{0.6\textwidth}\vspace{-35pt}
    \scriptsize
    \begin{tabular}{c|c}
    \!\!\!\includegraphics[width=0.28\textwidth]{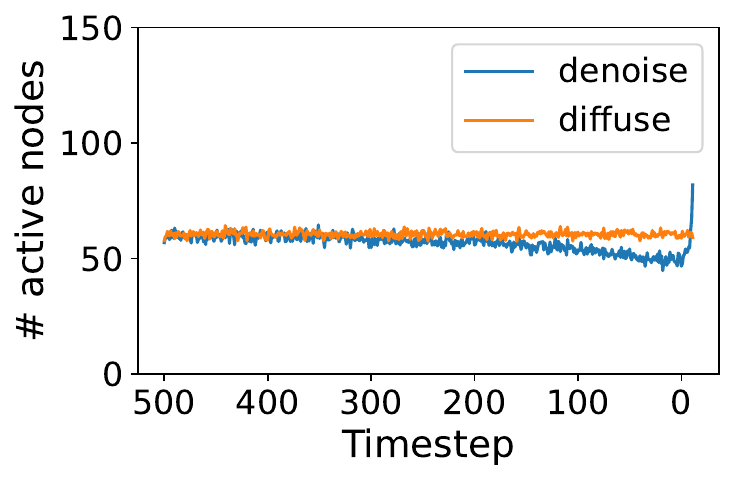} \!\!\!&\!\!\!\!
    \includegraphics[width=0.28\textwidth]{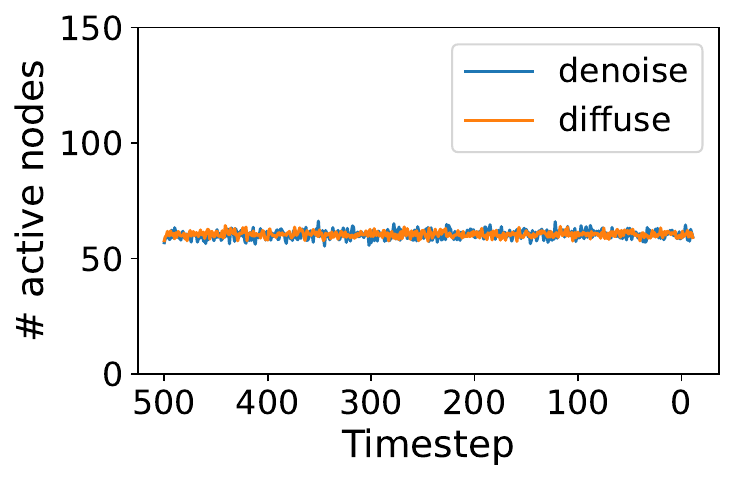}\\ 
    \!\!\!\includegraphics[width=0.28\textwidth]{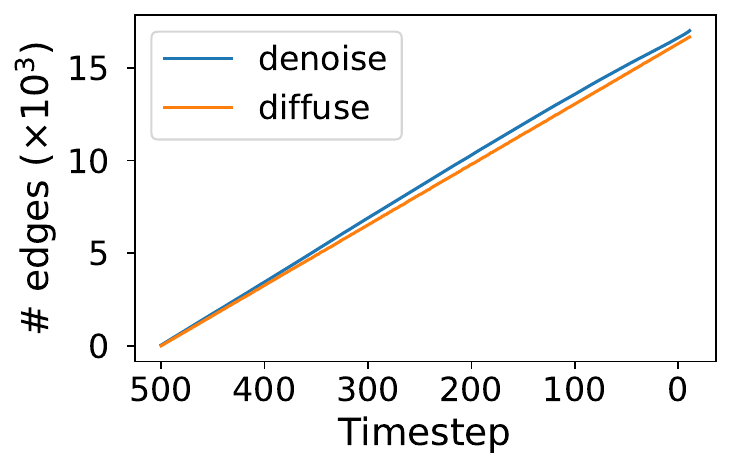} \!\!\!&\!\!\!\!
    \includegraphics[width=0.28\textwidth]{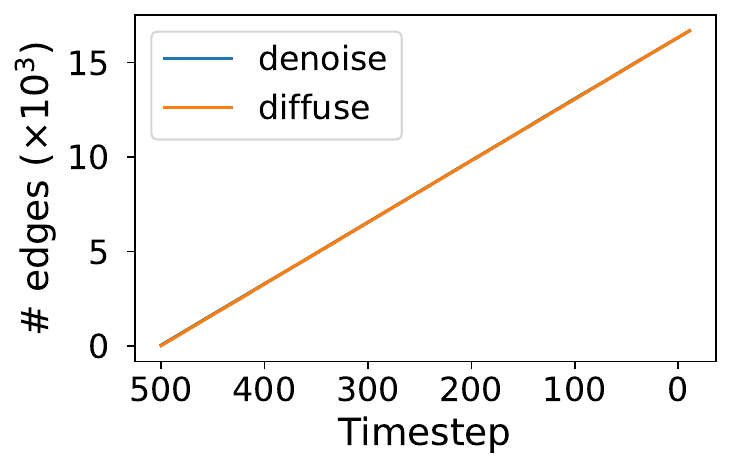} \\
        (a) Constant active node schedule & (b) Volume-preserved sampling
    \end{tabular}
    \setcounter{figure}{1}    
    \caption{Figure 2. \small{(a) Customized active node behavior still faces the mismatch problem; (b) Volume-preserved sampling corrects the mismatch problem.}}
    \label{fig:demo-fix}
\end{wrapfigure}

We identify that in EDGE, the use of degree-guided posterior $q(\bs|\bd^t,\bd^0)$ may lead to inconsistency between the active node behavior in diffusion and denoising processes. The reason is that when sampling edges from edge model $p_\theta(\bA^{t-1}|\bA^t,\bs^t)$, the degree constraints are not imposed. As a result, some nodes may stay inactive with more edges than their degree specification, while other nodes that are under the budget will still be sampled to be active. Such error can not be fixed by the degree-guided posterior and thus will accumulate over the denoising process.

\begin{figure}[t]
  \centering
  \begin{minipage}{0.54\textwidth} 
    \begin{algorithm}[H]
      \caption{Sovling $\alpha_{1:T}^*$ via binary search}
      \label{alg-obj}
      \small      
      \begin{algorithmic}[1]
        \State \textbf{Input}: Loss and constrainttolerance $\epsilon_1, \epsilon_2$, $K_\mathrm{min}, K_\mathrm{max}$, degree sequence $\bd^0$, and active node control $\gamma_{1:T}$.
        \State $K_1$ = $K_\mathrm{min}, K_2=K_\mathrm{max}, K=(K_1+K_2)/2$
        \State $\alpha_{1:T}^*=\argmin_{\alpha_{1:T}}\calL(\alpha_{1:T};K,\gamma_{1:T},\bd^0)$
        \While{$\calL(\alpha_{1:T}^*;\cdot)>\epsilon_1$ or $\prod_{t=1}^T\alpha_t^*>\epsilon_2$}
          \If{$\calL(\alpha_{1:T}^*;\cdot)>\epsilon_1$}
            \State $K_1=K, K=(K+K_2)/2$
          \ElsIf{$\prod_{t=1}^T\alpha_t^*>\epsilon_2$}
            \State $K_2 = K, K = (K+K_1)/2$
          \EndIf
        \State $\alpha_{1:T}^*=\argmin_{\alpha_{1:T}}\calL(\alpha_{1:T};K,\gamma_{1:T},\bd^0)$
        \EndWhile
        \State \textbf{Output}: Edge noise schedule parameter $\alpha_{1:T}^*$
      \end{algorithmic}
    \end{algorithm}
  \end{minipage}\hspace{5pt}
  \begin{minipage}{0.44\textwidth} 
    \begin{algorithm}[H]
    \small
      \caption{Volume-preserved sampling via reweighting}
      \label{alg-sampling}
      \begin{algorithmic}[1]
        \State \textbf{Input}: Degree sequence $\bd^0$, edge model $p_\theta$, denoising timestep $T$, and edge noise schedule parameter $\alpha_{1:T}$.
        \State Initialize $\bA^T = \bzero$
        \For{$t=T,\ldots,1$}
        \State Compute degree sequence $\bd^t$ from $\bA^t$
        \State Compute $\hat{q}(\bs^t|\bd^t,\bd^0)$ using Eq.~\ref{eq:node-reweight}
        \State Sample $\bs^t\sim\hat{q}(\bs^t|\bd^t,\bd^0)$
        \State Compute $\hat{p}_\theta(\bA^{t-1}|\bA^{t},\bs^{t})$ using Eq.~\ref{eq:edge-reweight}
        \State Sample $\bA^{t-1}\sim\hat{p}_\theta(\bA^{t-1}|\bA^{t},\bs^{t})$
        \EndFor
        \State \textbf{Output}: Generated graph $\bA^0$
      \end{algorithmic}
    \end{algorithm}
  \end{minipage}
  \vspace{-1em}
\end{figure}
We propose a solution that can guarantee the model generates the correct numbers of active nodes and edges for each timestep (See Fig.~\ref{fig:demo-fix}). This is achieved by simply reweighting the node and edge distribution. For active node distribution, we have the following corrected form:
\small
\begin{align}\hat{q}(\bs^t|\bd^t,\!\bd^0)\!=\!\prod_{i=1}^N\hat{q}(\bs_i^t|\bd^t,\!\bd^0),~\hat{q}(\bs_i^t|\bd^t,\bd^0)\!=\!\calB\bigg(\bs^t_i;\frac{h_{\bd^0}(\alpha_{1:t}, t)}{\sum_{i=1}^Np_i^t}p_i^t\bigg),~p_i^t\!=\!1-\Big(1-\frac{\beta_{t}\bar{\alpha}_{t-1}}{1\!-\!\bar{\alpha}_{t}}\Big)^{\bd^0_i-\bd^t_i}.\label{eq:node-reweight}
\end{align}
\normalsize
\normalsize
Recall that given edge noise schedule $\alpha_{1:T}$, $h_{\bd^0}(\alpha_{1:t}, t)$ is the expected number of active nodes at timestep $t$. The reweighting is performed on active nodes that still have a degree budget, i.e., $\bd_i^t<\bd_i^0$. For edge distribution, we have
\begin{align}
&\hat{p}_\theta(\bA^{t-1}|\bA^{t},\bs^{t},\bd^0)=\prod_{i,j:i<j}
\calB\big(\bA_{i,j}^{t-1};\frac{\Delta E_t}{\sum_{i^\prime,j^\prime:i^\prime<j^\prime}\ell_{\theta}^t(i^\prime,j^\prime)}\ell_{\theta}^t(i,j)\big),\label{eq:edge-reweight}\\
&\text{where}~\ell_\theta^t(i,j) = \hat{p}_\theta(\bA^{t-1}_{i,j}=1|\bA^{t},\bs^{t}),~\text{and}~\Delta E_t=\big((\bar{\alpha}_{t-1}-\bar{\alpha}_{t})\sum_{i=1}^N\bd_i^0+{\bs^t}^T\bA^t\bs^t\big)/2.
\nonumber
\end{align} 
Here $\ell_\theta^t(i,j)$ is the probability of forming an edge between node $i$ and $j$. Note that we only compute probabilities for active node pairs, and $\hat{p}_\theta(\bA^{t-1}_{i,j}=1|\bA^{t},\bs^{t})=\bA_{i,j}^{t}$ if one of the $\{i,j\}$ is inactive. Note that in each denoising step, we regenerate all edges within the subgraph indicated by $\bs^t$ as this allows the model to refine its previous prediction. The denominator of the weight in Eqn.~\ref{eq:edge-reweight} is the expected number of edges the model will generate within the subgraph. And the numerator $\Delta E_t$ represents the actual number of edges it should generate within the subgraph at that moment. With such reweighting, we can guarantee the model generates the correct number of edges at each timestep.

The correction operations on nodes and edges are optional and can be performed separately. We demonstrate the proposed sampling scheme in Alg.~\ref{alg-sampling}. With the corrected sampling algorithm, one can perform graph generation with edge overlap control~\citep{chanpuriya2021power}.

\vspace{-0.5em}
\section{Experiments}\vspace{-0.5em}
We demonstrate how the proposed techniques improve EDGE in terms of generative accuracy and efficiency. We denote the improved EDGE as EDGE++. We also present an application for generating realistic graphs with precise edge overlap control. 
\vspace{-0.5em}
\subsection{Setup}\vspace{-0.5em}

\paragraph{Datasets.} We perform experiments on two large networks: Polblogs~\citep{adamic2005political} and PPI~\citep{stark2010biogrid}. The Polblogs network contains 1,222 nodes and 16,714 edges in total, and the PPI network contains 3,852 nodes and 37,841 edges.
\vspace{-0.5em}
\paragraph{Evaluation.} We follow \citet{chanpuriya2021power} and \citet{chen2023efficient} to assess the consistency of the graph statistics between the generated networks and the original one. Our evaluation metrics encompass the following graph statistics: maximum degree; 
normalized triangle counts (NTC);
normalized square counts (NSC);
power-law exponent of the degree sequence (PLE);  
GINI;
assortativity coefficient (AC)~\citep{newman2002assortative}; 
global clustering coefficient (CC)~\citep{chanpuriya2021power}; 
and characteristic path length (CPL). We also access the memory consumption of the models during training and sampling.
\vspace{-0.5em}
\paragraph{Baselines.} Since EDGE has shown superior results over traditional baselines~\citep{rendsburg2020netgan, seshadhri2020impossibility, chanpuriya2021power}, we only compare to EDGE in \S~\ref{sec-4.2}. In \S~\ref{sec-4.3} We also compare against CELL~\citep{rendsburg2020netgan}, TSVD~\citep{seshadhri2020impossibility}, and three methods proposed by~\citet{chanpuriya2021power} (CCOP, HDOP, Linear).

\vspace{-0.5em}
\subsection{Generative Performance}
\label{sec-4.2}
We directly compare the generative performance of EDGE and EDGE++ in Table~\ref{tab:perf}. EDGE++ achieves competitive or better performance than EDGE in terms of recovering the graph statistics. Specifically, EDGE++ excels in 7 out of 8 metrics in both Polblogs and PPI datasets. \textbf{We hypothesize the reason is that the complexity of the edge prediction task is amortized to each timestep, the model is then more capable of a relatively simpler learning task.}  Moreover, training EDGE++ is more memory-economic compared to EDGE: it saves 31.25\% and 40.78\% of the GPU memory in training Polblogs and PPI datasets, respectively. This further demonstrates that with the proposed techniques, one can scale EDGE to model even larger graphs.
\begin{table}[t]
\centering
\small
\begin{tabular}{lcccccccccc}
\toprule
& \multicolumn{8}{c}{Graph statistics} & \multicolumn{2}{c}{Memory usage (GB)}\\
\!\!&\!\! Max Deg. \!\!&\!\! NTC \!\!&\!\! NSC \!\!&\!\! PLE \!\!&\!\! GINI \!\!&\!\! AC \!\!&\!\! CC \!\!&\!\! CPL \!\!&\!\! Training \!\!&\!\! Sampling \\
\midrule
\multicolumn{11}{c}{Polblogs}\\
\midrule
True \!\!&\!\! 351 \!\!&\!\! 1 \!\!&\!\! 1 \!\!&\!\! 1.414 \!\!&\!\! 0.622 \!\!&\!\! -0.221 \!\!&\!\! 0.23 \!\!&\!\! 2.738 \!\!&\!\! - \!\!&\!\! - \\
EDGE \!\!\!\!\!&\!\! \textbf{355.1} \!\!&\!\! \textbf{1.018} \!\!&\!\! 1.052 \!\!&\!\! \textbf{1.400} \!\!&\!\! \textbf{0.611} \!\!&\!\! -0.166 \!\!&\!\! 0.239 \!\!&\!\! 2.589 \!\!&\!\! 22.4 \!\!&\!\! 6.5 \\
EDGE++ \!\!&\!\! 344.2 \!\!&\!\! \textbf{1.016} \!\!&\!\! \textbf{1.023} \!\!&\!\! \textbf{1.401} \!\!&\!\! \textbf{0.603} \!\!&\!\! \textbf{-0.201} \!\!&\!\! \textbf{0.226} \!\!&\!\! \textbf{2.663} \!\!&\!\! \textbf{15.4} \!\!&\!\! \textbf{6.1} \\
\midrule
\multicolumn{11}{c}{PPI}\\
\midrule
True \!\!&\!\! 593 \!\!&\!\! 1 \!\!&\!\! 1 \!\!&\!\! 1.462 \!\!&\!\! 0.629 \!\!&\!\! -0.099 \!\!&\!\! 0.092 \!\!&\!\! 3.095 \!\!&\!\! - \!\!&\!\! - \\
EDGE \!\!&\!\! \textbf{593.5} \!\!&\!\! 1.143 \!\!&\!\! 1.601 \!\!&\!\! \textbf{1.431} \!\!&\!\! \textbf{0.604} \!\!&\!\! -0.062 \!\!&\!\! \textbf{0.102} \!\!&\!\! \textbf{3.071} \!\!&\!\! 57.6 \!\!&\!\! 19.5 \\
EDGE++ \!\!&\!\! \textbf{594.1} \!\!&\!\! \textbf{0.905} \!\!&\!\! \textbf{1.254} \!\!&\!\! \textbf{1.440} \!\!&\!\! \textbf{0.612} \!\!&\!\! \textbf{-0.081} \!\!&\!\! \textbf{0.082} \!\!&\!\! 3.011 \!\!&\!\! \textbf{34.1} \!\!&\!\! \textbf{17.2} \\
\bottomrule
\end{tabular}
\caption{Generative performance of EDGE and EDGE++. For graph statistics, we report the average statistics over 8 generated samples,  the numbers in bold indicate the
method is better at the 5\% significance level. Memory Usage of training is reported with a batch size of 4 and sampling with a batch size of 1.}
\label{tab:perf}\vspace{-1em}
\end{table}
\vspace{-0.5em}
\subsection{Graph Generation with EO Control -- An Application}
\begin{figure}[t]
    \centering
    \includegraphics[width=\textwidth]{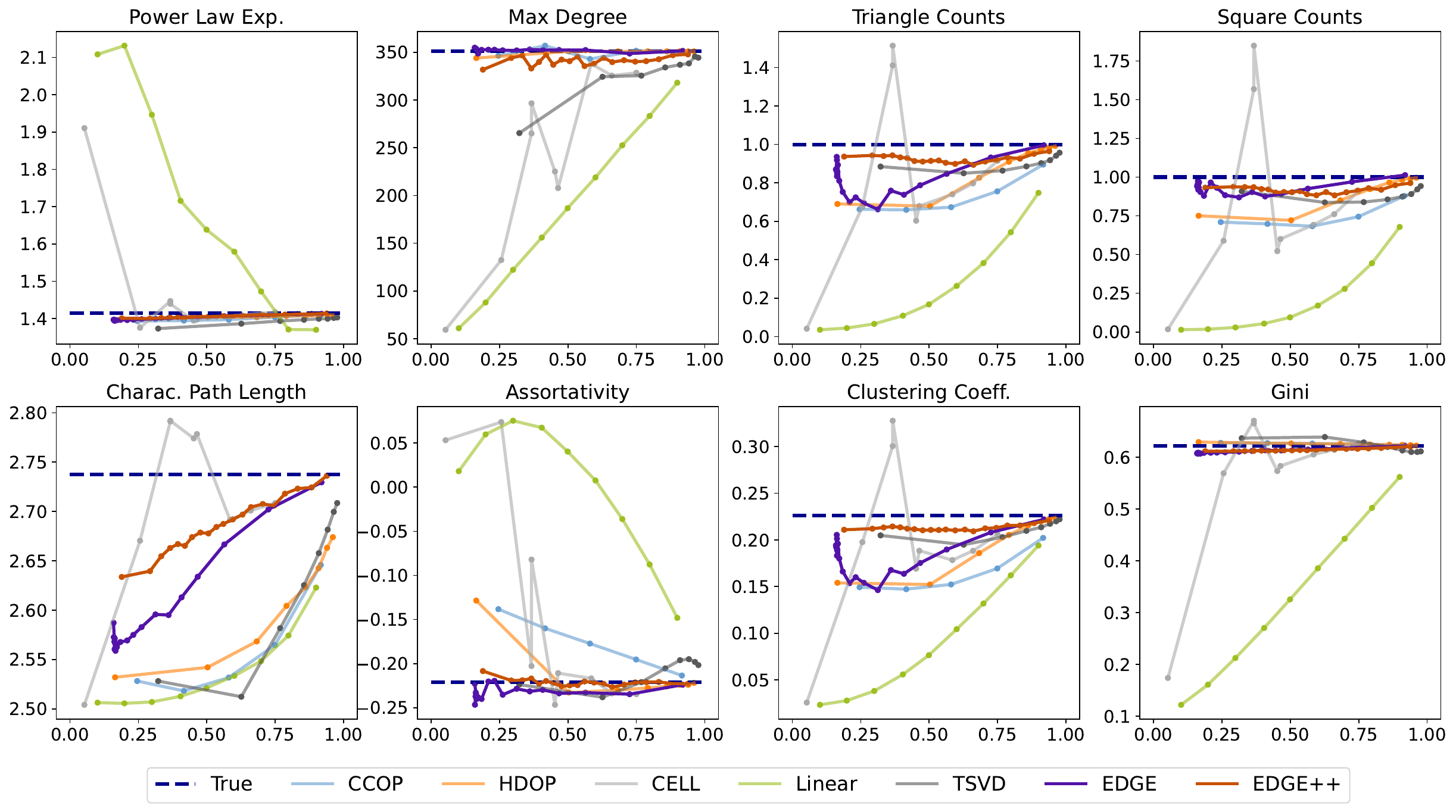}
    \caption{Figure 3. Graph generation of Polblogs with edge overlap control}
    \label{fig:eo-polblogs}\vspace{-0.5em}\vspace{-0.5em}
\end{figure}

\vspace{-0.5em}
\citet{chanpuriya2021power} shows that edge-independent graph models cannot generate desired graph statistics when the EO is low. Specifically, it tunes the EO to control the similarity between the generated graphs and the original graph. This section demonstrates how our proposed methods empower EDGE to generate graphs with controllable EOs. As we can observe in Fig.~\ref{fig:eo-polblogs}, the performance of EDGE degenerates when tunning EO from $0$ to $1$. However, such a phenomenon is not observed in EDGE++, indicating that we can synthesize realistic graphs with different levels of diversity. We further elaborate on how to control the EO of EDGE and EDGE++ in the App.~\ref{app:exp-eo}
\label{sec-4.3}
\vspace{-0.5em}\vspace{-0.5em}\section{Conclusion}\vspace{-0.5em}
In this work, we propose two techniques to improve the accuracy and efficiency of EDGE, a generative graph model that is able to generate high-quality large graphs. By customizing the number of active nodes at each timestep, EDGE requires significantly less GPU memory during training and achieves better generative performance on Polblogs and PPI databases. Moreover, the proposed volume-preserved sampling alleviates the trajectory mismatch problem, enabling one to generate graphs with edge overlap control. Our empirical study validates the effectiveness of the proposed techniques.

\bibliography{references}

\appendix
\section{Derivation of Active Node Control}
\label{app:derivation-connect}
We are interested in computing the expected number of active nodes given a prescribed degree sequence and an edge noise schedule $\alpha_{1:T}$:
\begin{align}
    h_{\bd^0}(\alpha_{1:t},t)= \sum_{i=1}^{N} q(\bs_i^t = 1| \bd_i^0),
\end{align}

where $q(\bs_i^t | \bd_i^0)$ is the distribution of node $i$ being active at timestep $t$. Before deriving the form of the distributions, we first revisit the two properties derived by~\citet{chen2023efficient}:

\textbf{Property 1.} The forward degree distributions have the form
\begin{align}
    &q(\bd^{t-1}|\bd^0)=\prod_{i=1}^N q(\bd^{t-1}_i|\bd^0_i),~\text{where} ~q(\bd^{t-1}_i|\bd^0_i)=\mathrm{Binomial}(k=\bd^{t-1}_i, n=\bd^0_i, p=\bar{\alpha}_{t-1}).
\end{align}
Intuitively, for $q(\bd^{t-1}|\bd^0)$, there are $\bd_i^0$ edges connected to node $i$, each with probability $\bar{\alpha}_{t-1}$ to be kept at time step ${t-1}$. The probability the number of remaining edges equals $\bd_i^{t-1}$ at time step ${t-1}$ is a binomial distribution. 

\textbf{Property 2.}  At timestep $t$, the active node distribution for node $i$ given $\bd^{t-1}_i$ is
\begin{align}
    q(\bs^t_i|\bd^{t-1}_i)=\calB\big(\bs^t_i;1-(1-\beta_t)^{\bd^{t-1}_i}\big). \label{eq:forward-deg-change}
\end{align}

With the above properties, we show that when $t=1$, we directly have
\begin{align}
    q(\bs_i^1 | \bd_i^0) = \calB(\bs_i^1 ; 1-(1-\beta_t)^{\bd_i^0}) 
    = 1-\alpha_t^{\bd_i^0}
\end{align}
by using property 1. For $t>1$, $q(\bs^t_i|\bd_i^0)$ can be computed by marginalization, specifically, we introduce $\bd_i^{t-1}$ and expand $q(\bs^t_i|\bd_i^0)$ into follow:
\begin{align}
    q(\bs_i^t | \bd_i^0)  
    &=  \sum_{\bd_i^{t-1}=1}^{\bd_i^0} q(\bs_i^t, \bd_i^{t-1} | \bd_i^0) \nonumber\\
    &=  \sum_{\bd_i^{t-1}=1}^{\bd_i^0} q(\bs_i^t | \bd_i^{t-1}) q(\bd_i^{t-1} | \bd_i^0).\nonumber\\
    &= \sum_{\bd_i^{t-1}=1}^{\bd_i^0} \calB(\bs_i^t ; 1-\alpha_t^{\bd_i^{t-1}}) \nonumber 
    \mathrm{Bin}(k=\bd_i^{t-1}, n=\bd_i^0, p=\bar{\alpha}_{t-1})\nonumber\\
    &=  \sum_{\bd_i^{t-1}=1}^{\bd_i^0}(1-\alpha_t^{\bd_i^{t-1}}) 
    \mathrm{Bin}(k=\bd_i^{t-1}, n=\bd_i^0, p=\bar{\alpha}_{t-1}).\nonumber
\end{align}

Now we show $q(\bs^t_i|\bd_i^0)$ as
\begin{align}
     q(\bs_i^t = 1| \bd_i^0) =
     \begin{cases}
        \displaystyle 1-\alpha_1^{\bd_i^0},~&\text{if}~t=1\\\\
        \displaystyle \sum_{\bd_i^{t-1}=1}^{\bd_i^0}(1-\alpha_t^{\bd_i^{t-1}}) \ 
        \mathrm{Bin}(k=\bd_i^{t-1}, n=\bd_i^0, p=\bar{\alpha}_{t-1}),&\text{otherwise}
     \end{cases}.
\end{align}

Then we can derive the expected number of active nodes as
\begin{align}
     h_{\bd^0}(\alpha_{1:t},t) =
     \begin{cases}
        \displaystyle\sum_{i=1}^{N} (1-\alpha_1^{\bd_i^0}),~&\text{if}~t=1\\\\
        \displaystyle\sum_{i=1}^{N} \sum_{\bd_i^{t-1}=1}^{\bd_i^0}(1-\alpha_t^{\bd_i^{t-1}}) \ 
        \mathrm{Bin}(k=\bd_i^{t-1}, n=\bd_i^0, p=\bar{\alpha}_{t-1}),&\text{otherwise}
     \end{cases}.
\end{align}

\section{Experiment Details}
\subsection{Experiment Setup}
For both Polblogs and PPI datasets, we set the number for diffusion timesteps to 512. We use the same architecture from~\citep{chen2023efficient}, with 5 message-passing blocks, each with 8 attention heads. 
We employ the Adam optimizer~\cite{kingma2014adam} with a weight decay of $10^{-4}$ and use a batch size of 4 for both datasets. The learning rate was fixed at $10^{-4}$, and we did not employ any learning rate scheduler during the training process. For model evaluation, we selected the model that minimized the statistics difference between the generated graphs and the original graphs. We use a batch size of 8 during evaluation.

\subsection{Graph Generation with EO control}
\label{app:exp-eo}
We can control the EO between the generated graphs and the original graph by generating $\bA^0$ from $\bA^t$ with different $\bA^t$. For any timestep $t\in [0,T]$, we can first draw $\bA^t\sim q(\bA^t|\bA^0)$, then use the learned denoising model to draw $\bA^0$ by sequentially drawing $(\bA^{0:t-1},\bs^{1:t})$ from $p_\theta(\bA^{0:t-1},\bs^{1:t}|\bA^t)$. In the experiment of \S~\ref{sec-4.3}, we choose $T=512$ and time interval $\Delta=25$, for each $t\in\{\Delta, 2\Delta,3\Delta,\ldots,T\}$, we generate eight graphs and evaluate their statistics.

For better visualization, we choose constant active node scheduling for EDGE++, which leads to that the density of $\bA^t$ decreases linearly as $t\rightarrow T$. Note that since EDGE doesn't support customization of the active node control, most of the sampled $\bA^0$ appear to have a very low EO. Moreover, due to the mismatching issue, EDGE conversely performs worse in terms of recovering the true statistics as EO increases.

\section{Related Works}
Edge-independent models, which assume the independent formation of edges with certain probabilities, are commonly found in probabilistic models for large networks. This category comprises various traditional models like the \erdosrenyi graph models~\citep{erdos1960evolution}, SBMs~\citep{holland1983stochastic}, and deep neural models like variational graph auto-encoders~\citep{kipf2016variational,mehta2019stochastic, li2020dirichlet, chen2022interpretable}, NetGAN and its variant~\citep{bojchevski2018netgan, rendsburg2020netgan}. Recent studies reveal that these models fail to replicate desired statistics of the target network, such as triangle counts, clustering coefficient, and square counts~\citep{seshadhri2020impossibility, chanpuriya2021power}.

On the other hand, deep auto-regressive (AR) graph models~\citep{li2018learning, you2018graphrnn, liao2019efficient, zang2020moflow, han2023fitting} construct graph edges by sequentially generating elements of an adjacency matrix. These algorithms are notably slow as they require making $N^2$ predictions. \citet{dai2020scalable} propose a method to circumvent this by leveraging graph sparsity and predicting only non-zero entries in the adjacency matrix. However, these AR-based models often face long-term memory issues, making it difficult to model global graph properties. These models also lack invariance with respect to node orders of training graphs, necessitating specialized techniques for their training\citep{chen2021order, han2023fitting}.

Diffusion-based generative models have been demonstrated to be effective in producing high-quality graphs~\citep{niu2020permutation, liu2019graph, jo2022score, haefeli2022diffusion, chen2022nvdiff, vignac2022digress,kongautoregressive,chen2023efficient}. They model edge correlations by refining a graph through multiple steps, overcoming the limitations of auto-regressive models. While the majority of the diffusion-based models have primarily focused on generation tasks with smaller graphs, EDGE~\citep{chen2023efficient} is the first model that scales to generate large graphs with thousands of nodes. However, EDGE is found to be inefficient in utilizing the denoising model. 

\begin{figure}[t]
\small
\centering
\begin{tabular}{ccc}
     \includegraphics[width=0.3\textwidth]{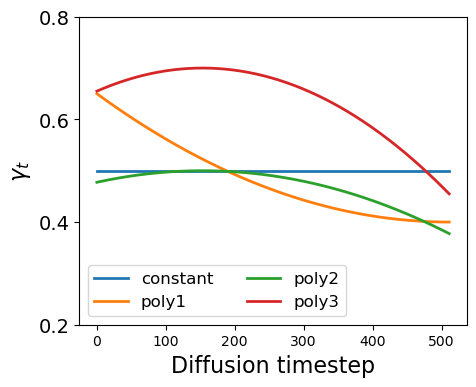} &  \includegraphics[width=0.3\textwidth]{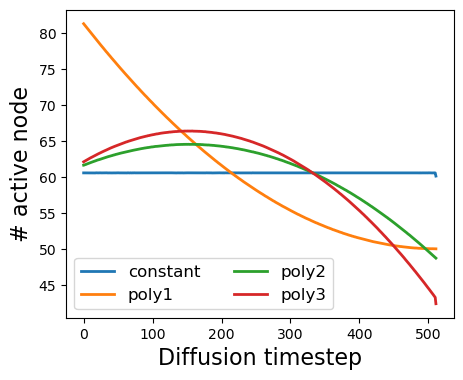} & \includegraphics[width=0.3\textwidth]{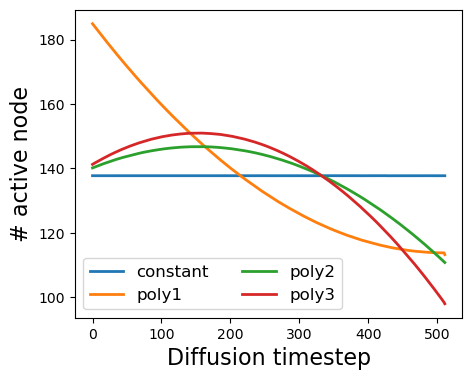} \\
     (a) schedule of $\gamma_{1:T}$& (b) active node schedule in Polblogs & (c) Active node schedule in PPI 
\end{tabular}
\caption{Figure 4. Visualization of how different $\gamma$ strategies may lead to different active node schedules on Polblogs and PPI datasets. }
\label{fig:vis-gamma-active}
\end{figure}
\section{Additional Results}
We investigate how to choose a suitable active node control for training EDGE. we found that constant schedule and polynomial schedule give better results in terms of generative performance. Specifically, we investigate the following three polynomial functions:
\begin{table}[H]
    \begin{tabular}{cl}
poly1: & $ \gamma_t = (0.5t/T - 0.5)^2 + 0.4 $ \\
poly2: & $ \gamma_t = (0.5t/T - 0.5)^2 + 0.5 $ \\
poly3: & $ \gamma_t = -0.5(t/T - 0.3)^2 + 0.7 $ 
    \end{tabular}
\end{table}

We visualize the chosen functions in Fig.~\ref{fig:vis-gamma-active}(a). Moreover, we also demonstrate how the $\gamma$ schedule maps to the actual node control in Polblogs (Fig.~\ref{fig:vis-gamma-active}(b)) and PPI  (Fig.~\ref{fig:vis-gamma-active}(c)). The active node schedule is data-specific since it needs to make sure the constraint $\prod\alpha_t \approx 0$ is satisfied. This is achieved by tuning the parameter $K$ as discussed in \S~\ref{sec:3.1}.

We further provide the generative performance of EDGE++ using different active node schedules in Table.~\ref{tab:abl}. All the active node schedules yield competitive performance in terms of recovering graph statistics. We report the result of the constant active node schedule in \S~\ref{sec-4.2}.
\begin{table}[h]
\centering
\small
\begin{tabular}{lcccccccc}
\toprule
& Max Deg. & NTC & NSC & PLE & GINI & AC & CC & CPL\\
\midrule
\multicolumn{9}{c}{Polblogs}\\
\midrule
True & 351 & 1 & 1 & 1.414 & 0.622 & -0.221 & 0.23 & 2.738\\
constant & 344.2 & 1.016 & 1.023 & 1.401 & 0.603 & -0.201 & 0.226 & 2.663\\
poly1 & 351.0 & 0.926 & 0.972 & 1.400 & 0.611& -0.171&0.205&2.634\\
poly2 & 352.3 & 0.922 & 0.913 & 1.400 & 0.613&-0.218&0.205&2.633\\
poly3 & 329.0 & 0.998 & 1.032 & 1.400 & 0.609 & -0.173&0.225&2.673\\
\midrule
\multicolumn{9}{c}{PPI}\\
\midrule
True & 593 & 1 & 1 & 1.462 & 0.629 & -0.099 & 0.092 & 3.095 \\
constant & 594.1 & 0.905 & 1.252 & 1.440 & 0.612 & -0.081 & 0.082 & 3.011  \\
poly1 & 594.3 & 0.904 & 1.244 & 1.440 & 0.612 & -0.081 & 0.088 & 3.029  \\
poly2 & 584.5 & 0.765 & 0.966 & 1.440 & 0.611 & -0.084 & 0.070 & 3.009  \\
poly3 & 594.0 & 0.743 & 0.999 & 1.442 & 0.614 & -0.099 & 0.059 & 3.026  \\
\bottomrule
\end{tabular}
\caption{Generative performance}
\label{tab:abl}
\end{table}
\end{document}